\documentclass[runningheads]{llncs}
\usepackage[T1]{fontenc}
\usepackage{graphicx}

\usepackage{wrapfig}
\usepackage{graphicx}
\usepackage{enumitem}
\usepackage{booktabs}
\usepackage{multirow}
\usepackage{makecell}
\usepackage[table]{xcolor}
\usepackage{boxedminipage}
\usepackage{tcolorbox}
\tcbuselibrary{breakable}
\definecolor{nmgray}{RGB}{229,229,229}
\definecolor{shadecolor}{RGB}{237,237,237}
\definecolor{darkpastelgreen}{rgb}{0.01, 0.75, 0.24}

\usepackage{pifont}
\newcommand{\xmark}{\ding{55}}
\usepackage{longtable}
\usepackage{adjustbox}
\usepackage{capt-of}
\usepackage{amsmath}
\usepackage{amssymb}

\begin{document}
\title{Exchange of Perspective Prompting Enhances Reasoning in Large Language Models}
%
%
\author{Lin Sun\orcidID{0009-0007-6960-3343} \and Can Zhang\orcidID{0000-0002-7083-5228}}
\authorrunning{Lin Sun}
%
\institute{UAES.AI, Shanghai, China\\
\email{\{lin.sun,can.zhang\}@uaes.com}}
\maketitle              
\begin{abstract}
    Large Language Models (LLMs) have made significant advancements in addressing 
    diverse natural language processing tasks. However, their performance is 
    often limited by inherent comprehension of problems. 
    To address this limitation, we propose Exchange-of-Perspective (EoP), a novel 
    framework designed to exchange perspectives across different definitions 
    of problem, so that it can break the fixed mindset from any particular formulation of 
    the question.
    We conducted extensive and comprehensive experiments on 8 benchmarks. The results show that EoP 
    can significantly improve performance. For instance, compared to the non-commutative 
    baseline PHP, with GPT-3.5-Turbo and EoP, we observe a {\bf +3.6\%} improvement on AQuA 
    (60.6\% → 64.2\%), while GPT-4-powered EoP demonstrates a {\bf +7.7\%} enhancement on Math 
    (53.9\% → 61.6\%) and a {\bf +3.5\%} improvement on OlympiadBench
    (43.5\% → 47.0\%) when using Qwen-2.5-72b.

\end{abstract}

\section{Introduction}

\begin{wrapfigure}{r}{0.5\textwidth}
    \vspace{-55pt}
    \centering
    \includegraphics[width=1.0\linewidth]{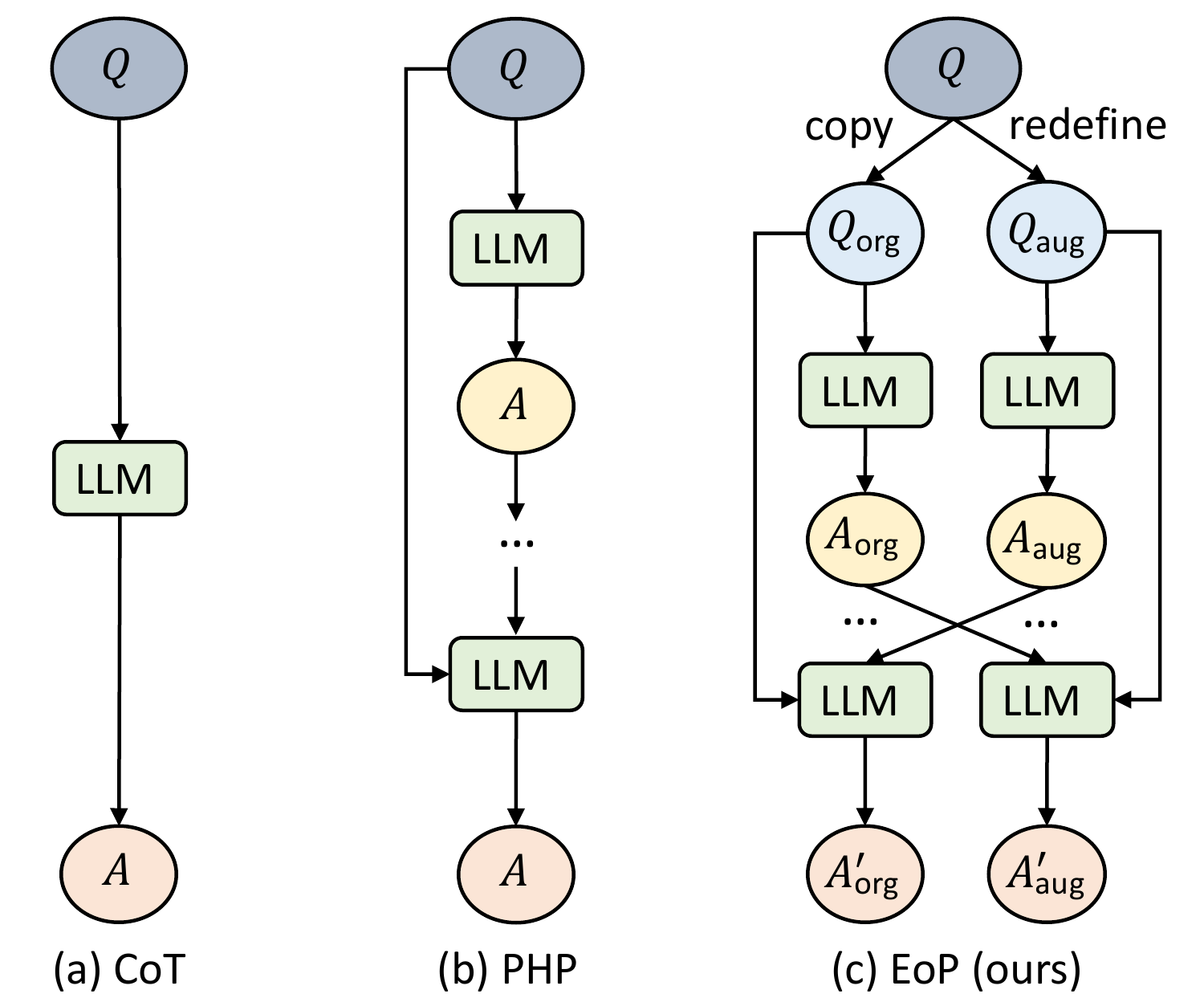}
    \caption{
    Comparison of CoT, PHP, and EoP. Both CoT and PHP rely on the model’s internal 
    perspective to generate or refine output, lacking external insights. 
     }
    \vspace{-15pt} 
    \label{fig:intro}
\end{wrapfigure}

LLMs have made significant progress in the field of NLP, but they often 
struggle to provide stable and accurate answers when faced with highly 
complex tasks. This issue cannot be resolved by simply scaling up the 
model size \cite{rae2021scaling,0002WSLCNCZ23}.

To address this limitation, chain-of-thought 
(CoT) prompting method was introduced \cite{wei2022chain}, which guides the model to generate a series of intermediate 
reasoning steps before arriving at the final answer. Subsequently, various 
self-correction strategies have emerged \cite{zheng2023progressivehintpromptingimprovesreasoning,welleck2023generating,ganguli2023capacity}. 
They are designed to iteratively 
improve the quality of responses by using the model's previous outputs.

Nevertheless, both CoT and self-correction techniques focus on reasoning
process, and they rely heavily on the model's 
own comprehension of the problem. Recent studies show that LLMs 
struggle to improve their responses without external feedback 
\cite{valmeekam2023largelanguagemodelsreally,stechly2023gpt4doesntknowits,0009CMZYSZ24}. This difficulty 
arises from their reliance on internal representations, making it challenging to overcome 
intrinsic capacity constraints \cite{YinSCGDHQ23}.

To tackle the challenges outlined, we propose a different viewpoint. Instead of centering 
focus on the reasoning process, {\bf let’s redirect our attention to the question}. 
We assert that it is preferable to thoroughly and deeply understand the question before 
formulating a solution rather than hastily offering a solution and then trying to revise it repeatedly.
One effective method of enhancing comprehension is to view the question from different perspectives, 
we observe a phenomenon that when humans answer questions, different definitions of the same question can lead 
to varied responses. This indicates that diverse phrasing of question has potential to yield multiple perspectives. 
When the responses are consistent, the likelihood of the response being correct also increases. 

\begin{figure}[!h]
    \centering
    \includegraphics[width=1.0\textwidth]{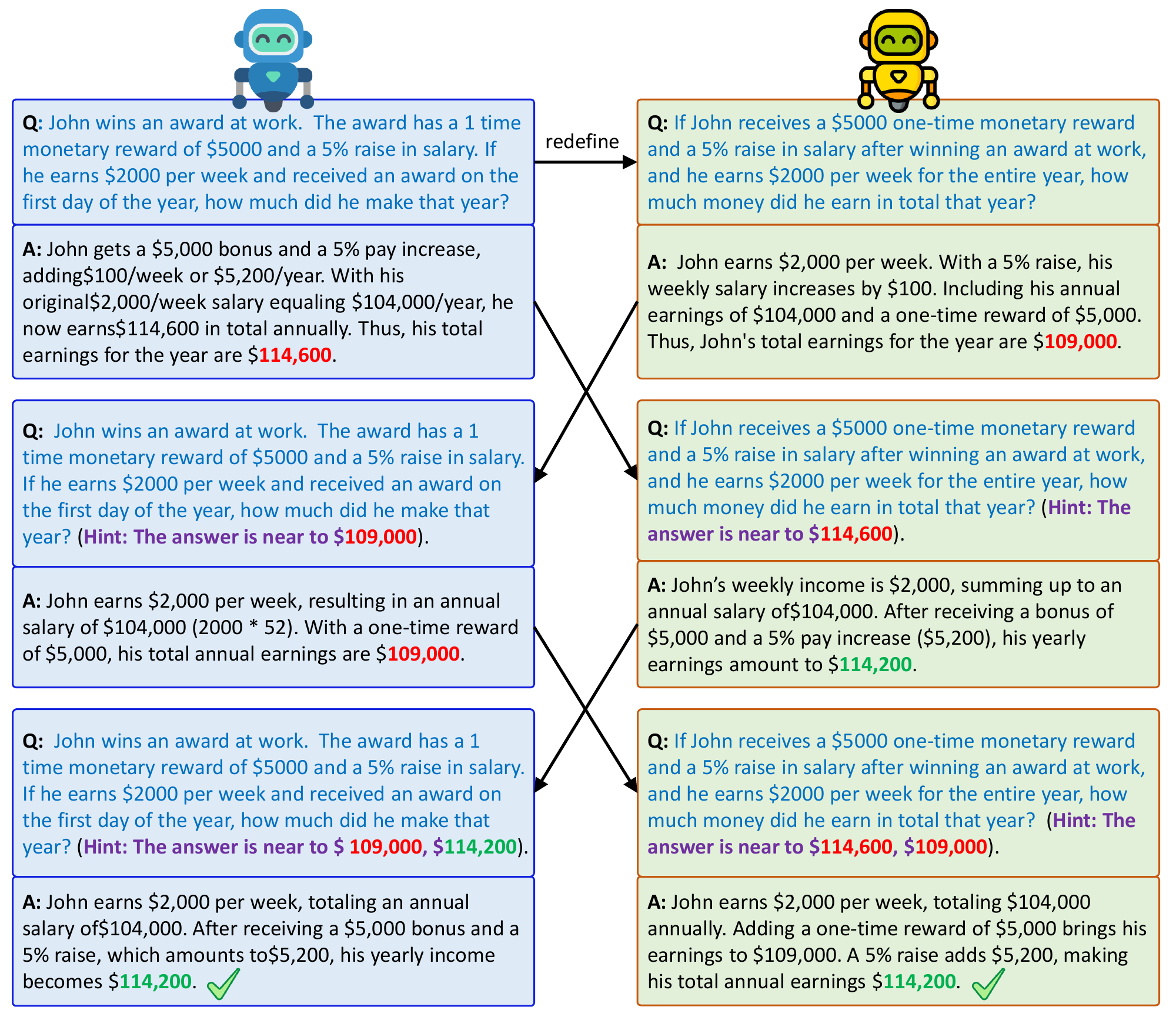}
    \caption{
    Our proposed EoP integrates the current question with answers from the alternative 
    branch to facilitate perspective exchange. It consists of four stages: 
    (1) We redefine the given question into an augmented version, subsequently forming 
    two branches: \textcolor{blue}{original branch} and \textcolor{orange}{augmented branch}. 
    (2) Instruct the LLM to produce initial answers for both the original and augmented 
    branches by providing it with a combination of the question and a fundamental prompt, 
    such as CoT or Complex CoT. 
    (3) Swap the answers and combine them with the question from the other 
    branch to generate subsequent answers.
    (4) We continue the process in step (3) until meeting termination condition. 
    }
    \label{fig:model_scheme}
    \end{figure}

Building on this insight, we introduce the Exchange-of-Perspective (EoP) framework, 
as shown in Figure \ref{fig:intro}. Unlike CoT and PHP \cite{zheng2023progressivehintpromptingimprovesreasoning}, 
EoP redefines the original question first, and then dynamically incorporates external perspectives by iteratively exchanging answers 
for the same question presented with different definitions.
Figure \ref{fig:model_scheme} illustrates EoP further. 
It executes as outlined below: 
(1) For a given question, we redefine it into an augmented question with LLM, subsequently forming 
two branches: the original branch and the augmented branch. 
(2) Instruct the LLM to generate initial answers for both branches. 
(3) Swap the answers and combine them with the question from 
the other branch using the phrase "Hint: The answer is near to", which follows 
\cite{zheng2023progressivehintpromptingimprovesreasoning},
to derive follow-up answers. 
(4) Continue the process in step (3) until meeting termination condition. 

We summarize our contributions as follows:  
\begin{itemize}[itemsep=2pt,topsep=0pt,parsep=0pt]  
    \item We introduce EoP, a novel framework that integrates various perspectives on the question, 
    {\bf we are exploring a new direction for improving LLM performance by focusing on the input side of 
    the question rather than the reasoning side}.
    \item We conduct extensive experiments across various complex reasoning tasks. 
    Results show that our method significantly outperforms established strong baselines, 
    highlighting the crucial role of external perspectives in enhancing the capabilities of LLMs.  
    \item Our research offers a unique viewpoint that yields a fresh perspective by 
    redefining the original problem. This method emphasizes the essential role of problem 
    definition in influencing comprehension and solutions.  
\end{itemize}

\section{Method}

\subsection{Architecture}
To deepen the understanding of question and acquire external perspective to enhance reasoning capability, the original question 
$\displaystyle q_\text{org}$ is passed through a redefinition function $\displaystyle f$, 
resulting in an augmented question, $\displaystyle q_\text{aug} = \displaystyle f(\displaystyle q_\text{org})$. 
The reasoning process then divides into two branches: the original branch and the augmented branch.

We denote the LLM with parameters $\theta$ as $\displaystyle P_{\theta}$. In the first interaction, 
the LLM produces a rationale $r^{(1)}_\text{org}$ and an answer $a^{(1)}_\text{org}$ 
for the original question $\displaystyle q_\text{org}$. Similarly, it generates a rationale 
$r^{(1)}_\text{aug}$ and an answer $a^{(1)}_\text{aug}$ for the augmented question 
$\displaystyle q_\text{aug}$. The first iteration of the reasoning process for both the original 
and augmented questions can be represented probabilistically as:

\begin{equation}
    (r^{(1)}_\text{org}, a^{(1)}_\text{org}) \sim \displaystyle P_\theta(r_\text{org}, a_\text{org} | \displaystyle q_\text{org}).
    \label{eq:p_org_initial}
\end{equation}

\begin{equation}
    (r^{(1)}_\text{aug}, a^{(1)}_\text{aug}) \sim \displaystyle P_\theta(r_\text{aug}, a_\text{aug} | \displaystyle q_\text{aug}).
    \label{eq:p_aug_initial}
\end{equation}

In subsequent iterations ($j > 1$), the model $P_\theta$ adapts its response strategy. 
It now generates a rationale $r^{(j)}_\text{org}$ and an answer $a^{(j)}_\text{org}$ 
for the original question $q_\text{org}$, taking into account the history 
of answers provided by the augmented question up to the $(j-1)$-th iteration, which 
is encapsulated in the set $A^{(j-1)}_\text{aug} = \{a^{(1)}_\text{aug}, ..., a^{(j-1)}_\text{aug}\}$. 
This process is modeled as:

\begin{equation}
    (r^{(j)}_\text{org}, a^{(j)}_\text{org}) \sim P_\theta(r_\text{org}, a_\text{org} | \displaystyle q_\text{org}, A^{(j-1)}_\text{aug}).
    \label{eq:p_org_follow}
\end{equation}

For augmented question $q_\text{aug}$, it uses $A^{(j-1)}_\text{org}= \{a^{(1)}_\text{org}, ..., a^{(j-1)}_\text{org}\}$ 
to generate $(r^{(j)}_\text{aug},a^{(j)}_\text{aug})$:

\begin{equation}
    (r^{(j)}_\text{aug}, a^{(j)}_\text{aug}) \sim P_\theta(r_\text{aug}, a_\text{aug} | \displaystyle q_\text{aug}, A^{(j-1)}_\text{org}).
    \label{eq:p_aug_follow}
\end{equation}

This cross-referencing between the original and augmented answers aims to acquire dynamic
external perspective to break the inherent fixed mindset of LLM, and enhance the model's 
ability to provide coherent responses, which can lead to more reliable and accurate 
results.

\subsection{Redefination Function}
The redefination of question adheres to two principles: (1) preserving the original 
semantics and (2) not changing the final answer. In line with these, we introduce 
two redefination strategies:

\paragraph{Premise Extraction and Concatenation (PEC).} 
Given original question
$q_\text{org}$, we pass it through LLM to extract key premises $[p_1, p_2, \dots, p_n]$ 
and the core question $q_\text{core}$. These 
elements are crucial for understanding the complex concepts presented in the question,
and then they are concatenated to form an augmented question $q_\text{aug}$:

\begin{equation}
    q_\text{aug} = [p_1, p_2, \dots, p_n] \oplus q_\text{core}.
    \label{eq:function_pec}
\end{equation}

The symbol $\oplus$ denotes concatenation. The main purpose of this method is to clarify 
the original question, making it clear and unambiguous.

\paragraph{Question Rephrasing (QR).} 
Utilize a LLM to directly rephrase the original question. This procedure 
involves first grasping the original question and subsequently expressing the 
question in a manner that aligns with the LLM’s understanding.

\subsection{Termination Conditions}
The iteration process terminates upon meeting one of the following conditions:

\paragraph{Consensus Across Branches.} 
For the $j$-th iteration, if the output from 
the original branch matches the augmented branch, i.e., 
$a^{(j)}_{\text{org}} = a^{(j)}_{\text{aug}}$, it indicates that both branches 
have reached a consensus.

\paragraph{Stability Within Branch.} 
For a given branch, if the output in the 
$j$-th iteration is identical to that of the $(j-1)$-th iteration, either 
$a^{(j)}_{\text{org}} = a^{(j-1)}_{\text{org}}$ for the original branch or 
$a^{(j)}_{\text{aug}} = a^{(j-1)}_{\text{aug}}$ for the augmented branch, 
it indicates that the branch's output is stable.

\section{Experiment}

\subsection{Experimental Settings}

\paragraph{Datasets and Models.} 
We evaluate the performance of EoP on 8 datasets: 
AddSub\cite{hosseini2014learning}, MultiArith \cite{roy-roth-2015-solving}, 
 SingleEQ \cite{koncel2015parsing}, SVAMP \cite{patel2021nlp}, 
 GSM8K \cite{abs-2110-14168}, AQuA \cite{ling2017program} and 
 Math \cite{HendrycksBKABTS21}, 
 OlympiadBench \cite{HeLBHTSHHHZLQL024}. 
 These datasets were selected to focus on
 model’s mathematical reasoning capability. We utilized three types of prompts: Standard, 
 Chain-of-Thought (CoT) \cite{wei2022chain}, and Complex CoT \cite{fu2022complexity}. 
 For more details, see Section \ref{sec:prompt}.
To verify the effectiveness of our proposed method, we employ 4 models: GPT-3.5-Turbo, GPT-4 
 \cite{ouyang2022training,openai2023gpt4} and Qwen-2.5-7b, Qwen-2.5-72b \cite{abs-2407-10671}.

\paragraph{Baselines.} 
We benchmark our proposed EoP against several strong 
baselines. For the Arithmetic dataset, the baselines include: 
(1) Chain-of-Thought prompting (CoT; \cite{kojima2022large}), 
(2) Plan-and-Solve prompting (PS; \cite{wang2023planandsolve}), 
(3) Least-to-Most prompting (\cite{zhou2023leasttomost}), 
(4) Contrastive Prompting (CP; CoT-CP; \cite{abs-2403-08211}), 
(5) Progressive-Hint Prompting (PHP, \cite{zheng2023progressivehintpromptingimprovesreasoning}).
Regarding the Math dataset, in addition to CoT and PHP, we also compared with
following methods:
(1) Program-Aided Language models (PAL; \cite{gao2023pal}),
(2) Tool-Integrated Reasoning Agent (ToRA; \cite{GouSGSYHDC24}), 
(3) Skills-in-Context Prompting (SKiC; \cite{abs-2308-00304}),
(4) Cumulative Reasoning (CR; \cite{abs-2308-04371}).
Additionally, we compared EoP with self-consistency (SC; \cite{0002WSLCNCZ23}) on 
the Math and OlympiadBench datasets, and we set the temperature 
$T = 0.8$ for SC method, and $T = 0$ for other baselines during testing.

\subsection{Results}

\paragraph{Performance on Arithmetic dataset.} 
Table \ref{tab:mainresults_arithmetic} displays the results of existing baselines 
and our EoP approach on the Arithmetic dataset. EoP achieves the highest mean accuracy, 
with GPT-3.5-Turbo recording 85.3\%, a 4.4\% improvement over the CoT baseline and 
surpassing PHP by 1.1\%. The AQuA dataset shows the most significant enhancement, 
with EoP scoring 64.2\%, reflecting an 11.2\% increase over CoT and a 3.6\% rise over 
PHP. For GPT-4, all baselines show improved performance, with EoP maintaining the 
highest mean accuracy. These results demonstrate that EoP is a robust and efficient 
method across various LLMs and datasets, particularly excelling in complex reasoning 
tasks like AQuA, highlighting its suitability for deep understanding tasks.


\begin{table*}[!h]
    \caption{
        {\bf Evaluation Results on Arithmetic Dataset:} 
        When applied to various LLMs with complex CoT prompts and PEC redefinition, 
        EoP outperforms baseline methods. {\bf Avg.} indicates mean accuracy across all 
        test datasets. Top results are in {\bf bold}, and runner-up results are 
        \underline{underlined}. Performance improvements ($\Delta$) are relative to each 
        baseline method.
    }
    \label{tab:mainresults_arithmetic}
    \begin{center}
    \setlength{\tabcolsep}{0.8mm}\small
    {
    \begin{tabular}{lcccccccc}
    
        \toprule
        \multirow{2}{*}{\bf Method} & \multicolumn{6}{c}{\bf Arithmetic Dataset} & \multirow{2}{*}{\bf Avg.}&\multirow{2}{*}{\bf $\Delta$}\\
        \cmidrule(r){2-7}
        & AddSub &  MultiArith & SingleEQ & SVAMP &  GSM8K & AQuA\\
        \midrule

        \rowcolor{gray!20}
        \multicolumn{9}{c}{\bf ChatGPT (\texttt{GPT-3.5-Turbo})} \\
        \midrule
        CoT & 85.8	& 95.3	& \underline{93.5}	& 79.3	& 78.9	& 53.0	& 80.9 &\textcolor{blue}{+4.4} \\
        PS & 86.6	& 93.8	& 92.5	& 79.4	& 76.1	& 58.9	& 81.2 &\textcolor{blue}{+4.1}\\
        Least-to-Most	& \textbf{91.3}	& 95.5	& \underline{93.5}	& 80.9	& 77.5	& 57.4	& 82.6 &\textcolor{blue}{+2.7}\\
        CP  & \underline{90.6}	& 95.2	& 91.7	& 83.2	& 73.2	& 40.2 & 80.8 &\textcolor{blue}{+4.5} \\
        CoT-CP  & 88.6	& 96.2	& 92.3	&\textbf{85.9}	& 73.5	& 60.6	& 82.9 &\textcolor{blue}{+2.4} \\
        PHP  & 85.3	& \underline{98.0}	& 92.9	& 83.1	& \textbf{85.1}	& 60.6	& \underline{84.2} &\textcolor{blue}{+1.1}\\
        EoP (ours) & 87.3	& \textbf{98.2}	& \textbf{93.6}	& \underline{84.6}	& \underline{84.2}	& \textbf{64.2}	& \textbf{85.3}\\
        \midrule
    
        \rowcolor{gray!20}
        \multicolumn{9}{c}{\bf GPT-4} \\
        \midrule
        CoT 	& 92.4	& 97.8	& 95	& 90.4	& 94.6	& 72.8	& 90.6 &\textcolor{blue}{+1.7}       \\
        PS 	& \underline{93.1}	& \underline{98.1}	& \textbf{95.3}	& \underline{92.6}	& 94.3	& 75.5	& \underline{91.4}   &\textcolor{blue}{+0.9}     \\
        Least-to-Most	& 92.1	& 97.1	& \underline{95.0}	& 90.3	& 92.1	& 71.6	& 89.7 &\textcolor{blue}{+2.6}\\
        CP &91.6	&97.8	&91.7	&91.5	&88.8	&62.2 & 87.3 &\textcolor{blue}{+5.0} \\
        CoT-CP  & 91.4	& 97.2	& 92.7	& 91.6	& 89.5	& 71.3	& 89.0 &\textcolor{blue}{+3.3}\\
        PHP 	& 89.6	& \underline{98.1}	& 93.1	& 91.9	& \underline{95.5}	& \textbf{79.9}	&  \underline{91.4} &\textcolor{blue}{+0.9}\\
        EoP (ours)	& \textbf{93.4}	& \textbf{98.3}	& 94.5	& \textbf{93.0}	& \textbf{96.2}	& \underline{78.4}	& \textbf{92.3} \\
        \bottomrule
    
    \end{tabular}
    }
    
    \end{center}
    \end{table*}


\begin{table*}[!h]
    \caption{{\bf Evaluation Results on Math Dataset:} 
    Significant improvements are observed for EoP technique on Math dataset.
    In addition to overall accuracy (\%), we provide a breakdown of accuracy
    for various question types within the test set. Both PHP and EoP are applied with complex CoT prompt.
    The results are from GPT-4 with greedy decoding and PEC redefination.
    }
    \label{tab:mainresults_math}
    \begin{center}
    \noindent\resizebox{\textwidth}{!}
    {   
        \small
        \begin{tabular}{lccccccccc}
    
        \toprule
        \multirow{2}{*}{\bf Method} & \multicolumn{7}{c}{\bf Math Dataset} & \multirow{2}{*}{\bf Avg.}&\multirow{2}{*}{\bf $\Delta$}\\
        \cmidrule(r){2-8}
        & Algebra	& Probability	& Geometry 	& InterAlgebra	& NumTheory	& PreAlgebra	& Precalculus\\
        \midrule
        CoT            & 70.8	& 53.1	& 36.5	& 23.4	& 49.6	& 71.6	& 26.7	& 50.3 &\textcolor{blue}{+11.3}\\
        PAL            & 59.1	& 61.0	& 38	& 32.8	& 58.7	& 73.9	& 29.3	& 52.0 &\textcolor{blue}{+9.6}\\
        ToRA           & 71.8	& \underline{66.1}	& \textbf{48.8}	& \textbf{49.5}	& 49.5	& 67.1	& \textbf{44.6}	& \underline{60.8} &\textcolor{blue}{+0.8}\\
        SKiC           & 74.6	& 58.2	& 43.6	& 29.5	& 55.9	& \underline{79.7}	& 36.6	& 56.4 &\textcolor{blue}{+5.2}\\
        CR             & \textbf{86.6}	& 63.2	& 43.9	& 32	& \underline{59.7}	& 71.8	& 35.7	& 58.0 &\textcolor{blue}{+3.6}\\
        PHP            & 74.3	& 56.3	& 41.9	& 26.3	& 55.7	& 73.8	& 29.8	& 53.9 &\textcolor{blue}{+7.7}\\
        EoP (ours)	   & \underline{80.1}	& \textbf{70.0}	& \underline{47.6}	& \underline{35.2}	& \textbf{63.5}	& \textbf{81.2}	& \underline{36.8}	& \textbf{61.6} \\
        \bottomrule
        
    \end{tabular}
    }
    
    \end{center}
    \vspace{-0.1in}
    \end{table*}

\paragraph{Performance on Math and OlympiadBench Maths dataset.} 
Table \ref{tab:math_and_OlympiadBench} presents the results on the Math and OlympiadBench 
Maths datasets using the Qwen-2.5 series. EoP achieves the best performance on these challenging 
tasks. Table \ref{tab:mainresults_math} further provides a breakdown of each question 
type on the Math dataset using GPT-4.
This time we observe even more improvements for the EoP approach. It is worth 
noting that PHP can be seen as a special case of EoP, focusing solely on the original 
branch. The results reveal that the EoP approach gets a mean accuracy of 61.6\%, 
surpassing PHP by 7.7\%.  Significantly, the EoP technique 
outperforms PAL and ToRA, which are code-based methods requiring the implementation 
of specific code. The consistent and reliable results across various mathematical 
problems highlight the method's robustness and its potential for enhancing reasoning 
capabilities.

\subsection{Ablation Study}

\begin{wrapfigure}{r}{0.5\textwidth} 
    \captionof{table}{
        {\bf Evaluation Results on Math and OlympiadBench Maths:} 
        N represents the average interaction number required to obtain the 
        final answer from the LLM in the reasoning phase.
    }
    \vspace{2pt} 
    \begin{adjustbox}{max width=\linewidth}
        \begin{tabular}{@{}l l c c c c@{}}
            \toprule
            \multirow{2}{*}{\bf LLM} & \multirow{2}{*}{\bf Method} & \multicolumn{2}{c}{\bfseries Math} & \multicolumn{2}{c}{\bfseries Olympiad} \\
            \cmidrule(lr){3-4} \cmidrule(lr){5-6}
            &  & \multicolumn{1}{c}{N} & \multicolumn{1}{c}{Acc.} & \multicolumn{1}{c}{N} & \multicolumn{1}{c}{Acc.} \\
            \midrule
            \multirow{4}{*}{Qwen2.5-7b} & CoT & 1.0 & 71.1 & 1.0 & 35.8 \\
                & PHP & 2.4 & \underline{72.5} & 2.5 & \underline{38.1} \\
                & EoP (ours) & 3.2 & {\bf 74.6} & 4.8 & {\bf 40.7} \\
                & SC & 4.0 & 71.8 & 5.0 & 37.2 \\
            \midrule
            \multirow{4}{*}{Qwen2.5-72b} & CoT & 1.0 & 78.5 & 1.0 & 42.1 \\
                & PHP & 2.3 & 79.2 & 2.4 & \underline{43.5} \\
                & EoP (ours) & 2.9 & {\bf 81.7} & 4.2 & {\bf 47.0} \\
                & SC & 4.0 & \underline{80.5} & 5.0 & 43.1 \\
            \bottomrule
        \end{tabular}
    \end{adjustbox}
    \label{tab:math_and_OlympiadBench}
    \vspace{-10pt} 
\end{wrapfigure}

\paragraph{Why EoP can get significant performance improvement?}
Table \ref{tab:ablation_study} showcases the performance metrics for each branch. 
The Org Branch (with original question) yields 83.3\% for CoT Prompt and 83.1\% for Complex CoT Prompt, 
while the Aug Branch (with rephrased questions) results in 82.1\% and 81.7\% 
respectively. These results indicate that the rephrased questions perform
even worse than the original questions. However, when integrating both branches, EoP framework outperforms 
individual branches with scores of 84.9\% for CoT Prompt and 85.3\% for Complex CoT 
Prompt. So the performance gain of EoP is not from rephrasing the question, while it comes from two main factors: (1) Error Correction, where insights 
from one branch can rectify misinterpretations from another, thereby improving problem 
analysis accuracy, and (2) Complementary Information, where merging branches provides 
more extensive and holistic insights to problem-solving.
So EoP can effectively mitigating the potential inaccuracies introduced by rephrased question.

\paragraph{PEC redefination method performs better.} 
Table \ref{tab:experiment_redefination} illustrates that the PEC method achieves 
a 1.3\% enhancement over QR using Standard prompt. Additionally, PEC exceeds QR 
by 0.7\% with CoT prompt, and demonstrates a 2.1\% advantage with complex CoT prompt. 
These results indicate that PEC consistently outperforms QR across various prompt
scenarios. The primary strength of the PEC approach lies in its capacity to clarify 
premises and the foundational question. This leads to more targeted and clear inquiries. 
Such precision is particularly beneficial for models tackling complex tasks, such as 
AddSub and AQuA, where a thorough understanding of the problem is crucial for achieving 
superior accuracy. While QR employs LLM to rephrase questions directly, it may enhance 
the limited inherent comprehension of the question and finally diminish performance.

\begin{table*}[!h]
    \caption{Ablation Study. We
    employ the prompt of CoT and Complex CoT. 
    {\bf Org}: original branch, 
    {\bf Aug}: augmented branch. 
    According to the experiment results, we see that the performance of combined branch exceeds 
    that of the individual branch across various tasks.
    The results are from GPT-3.5-Turbo with greedy decoding and PEC redefination.
    }
    \label{tab:ablation_study}
    \begin{center}
        \noindent\resizebox{\textwidth}{!}
    { 
        \small
        \begin{tabular}{cccccccccc}
            \toprule
                \multirow{2}{*}{\bf Prompt} & \multirow{2}{*}{\bf Org} & \multirow{2}{*}{\bf Aug} &\multicolumn{6}{c}{\bf Arithmetic Dataset}  &\multirow{2}{*}{\bf Avg.} 
                \\
                \cmidrule(r){4-9}
            &  & &AddSub&MultiArith &SingleEQ&SVAMP&GSM8K&AQuA\\
            \cmidrule(r){1-10}
            
            \multirow{3}{*}{CoT}     
            & \checkmark      &\xmark         &89.4	 &97.3	&93.9	&79.7	&78.6	&60.6	&83.3\\
            & \xmark          &\checkmark     &90.1	 &93.8	&93.3	&79.3	&75.7	&60.2	&82.1\\
            & \checkmark      &\checkmark     &\textbf{90.4}	 &\textbf{97.7}	&\textbf{94.5}	&\textbf{82.1}	&\textbf{80.4}	&\textbf{64.2}	&\textbf{84.9}\\
            \midrule
        
            \multirow{3}{*}{Complex CoT}     
            & \checkmark      &\xmark         &86.3	 &98.1	&93.3	&81.3	&81.4	&58.3	&83.1\\
            & \xmark          &\checkmark     &86.3	 &94.2	&92.3	&80.7	&79.5	&57.5	&81.7\\                 
            & \checkmark      &\checkmark     &\textbf{87.3}  &\textbf{98.2}	&\textbf{93.6}	&\textbf{84.6}	&\textbf{84.2}	&\textbf{64.2}	&\textbf{85.3}\\
            \bottomrule
        
    \end{tabular}
    }
    
    \end{center}
    \vspace{-0.1in}
    \end{table*}

\begin{figure}[!ht]
    \centering
    \includegraphics[width=1.0\textwidth]{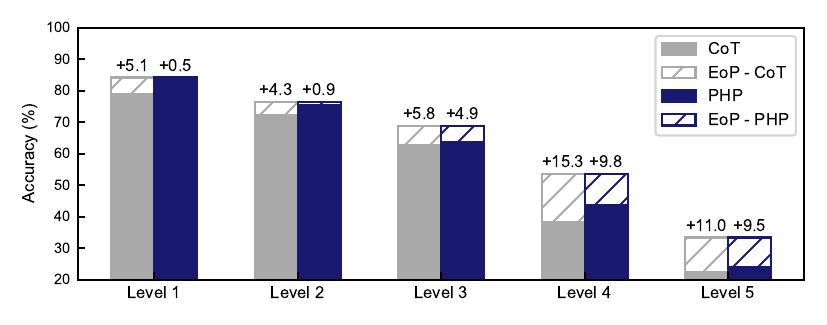}
    \caption{Performance comparison between CoT, PHP and EoP on math dataset with 
    varying difficulty levels. It shows that EoP achieves the best performance across 
    all difficulty levels. The enhancement in performance becomes more evident when 
    tackling more challenging problems. The results are based on GPT-4 with 
    greedy decoding and PEC redefination.}
    \label{fig:level_math}
    \end{figure}

\paragraph{EoP performs better when addressing more difficult challenges.}
The data illustrated in Figure \ref{fig:level_math} indicates that EoP exhibits 
significant enhancements when tackling more complex challenges, surpassing both 
CoT and PHP. Initially, all methods demonstrate comparable performance. However, 
as the difficulty of the 
problems increases, EoP's decline in performance becomes less evident, especially 
from Level 3 onwards. By Level 5, EoP achieves a score considerably higher than the 
other methods, boasting a 9.5\% advantage over PHP and an 11.0\% lead over CoT. 
This performance underscores EoP’s exceptional ability to handle intricate 
problems, primarily due to the collaborative and adaptive characteristics of the 
insight exchange process.

\begin{table*}[!h]
    \caption{Comparison of Question Redefination Methods.
    \textcolor{blue}{Blue}: The performance of PEC is better than that of QR. 
    \textcolor{red}{Red}: The performance of PEC is worse than that of QR. 
    According to the experiment results, we see that PEC redefination method performs better in most cases.
    The results are from GPT-3.5-Turbo with greedy decoding.}
    \label{tab:experiment_redefination}
    \begin{center}
    \noindent\resizebox{\textwidth}{!}
    {
        \small
        \begin{tabular}{ccccccccc}
            \toprule
             \multirow{2}{*}{\bf Prompt} & \multirow{2}{*}{\parbox{1.5cm}{\centering \textbf{Redefination }\\\textbf{Method}}} &\multicolumn{6}{c}{\bf Arithmetic Dataset}  &\multirow{2}{*}{\bf Avg.} 
             \\
             \cmidrule(r){3-8}
            &  &AddSub&MultiArith &SingleEQ&SVAMP&GSM8K&AQuA\\
            \midrule
            
            \multirow{3}{*}{Standard}     
            & \multirow{1}{*}{QR}          &85.6	&\textbf{87.5}	&90.2	&\textbf{81.0}	&57.0	&36.2 &72.9\\
            & \multirow{1}{*}{PEC}         &\textbf{90.9}	&85.2	&\textbf{90.7}	&80.2	&\textbf{57.2}	&\textbf{40.9} &\textbf{74.2}\\
            & \multirow{1}{*}{$\Delta$ Absolute gain}    &\textcolor{blue}{(+5.3)}	&\textcolor{red}{(-2.3)}	&\textcolor{blue}{(+0.5)}	&\textcolor{red}{(-0.8)}	&\textcolor{blue}{(+0.2)}	&\textcolor{blue}{(+4.7)} &\textcolor{blue}{(+1.3)}\\
            \midrule

            \multirow{3}{*}{CoT}     
            & \multirow{1}{*}{QR}          &87.3	&\textbf{98.7}	&93.1	&81.0	&\textbf{81.2}	&63.8 &84.2\\
            & \multirow{1}{*}{PEC}         &\textbf{90.4}	&97.7	&\textbf{94.5}	&\textbf{82.1}	&80.4	&\textbf{64.2} &\textbf{84.9}\\
            & \multirow{1}{*}{$\Delta$ Absolute gain}    &\textcolor{blue}{(+3.1)}	&\textcolor{red}{(-1.0)}	&\textcolor{blue}{(+1.4)}	&\textcolor{blue}{(+1.1)}	&\textcolor{red}{(-0.8)}	&\textcolor{blue}{(+0.4)} &\textcolor{blue}{(+0.7)}\\
            \midrule
      
            \multirow{3}{*}{Complex CoT}     
            & \multirow{1}{*}{QR}       &85.3	&\textbf{98.0}	&93.3	&\textbf{82.9}	&81.9	&57.1 &83.2\\
            & \multirow{1}{*}{PEC}      & \textbf{87.3}	&98.2	&\textbf{93.6}	&84.6	&\textbf{84.2}	&\textbf{64.2} &\textbf{85.3}\\
            & \multirow{1}{*}{$\Delta$ Absolute gain} &\textcolor{blue}{(+2.0)}	&\textcolor{blue}{(+0.2)}	&\textcolor{blue}{(+0.3)}	&\textcolor{blue}{(+1.7)}	&\textcolor{blue}{(+2.3)}	&\textcolor{blue}{(+7.1)} &\textcolor{blue}{(+2.1)}\\
            \bottomrule
        
    \end{tabular}
    }
    
    \end{center}
    \vspace{-0.1in}
    \end{table*}

\section{Related work}

\paragraph{Chain-of-Thought prompting.} 
Since CoT underscores the value of multi-step logical pathways in deriving conclusive 
answers, a series of enhancement strategies has been proposed. 
One noteworthy strategy is the careful selection of demonstrations. For instance, PromptPG
\cite{Lu0CWZRCK23} employs policy gradients to choose demonstrations. Research conducted 
by \cite{WangYW24} focused on training 
dense retrievers to select high-quality demonstrations. For complex tasks, a two-stage method 
is recommended, involving the breakdown of tasks into manageable sub-tasks that are solved 
sequentially before integration \cite{zhou2023leasttomost,KhotTFF0CS23}. 
Additionally, approaches such as Program-of-Thought (PoT; \cite{ChenM0C23}), 
Program-Aided Language models (PAL; \cite{gao2023pal}) propose to generate intermediate reasoning 
programs and employ external interpreters for execution. Moreover, equipping LLMs with external 
tools, such as scratch pads for intermediate computations, search engines for information retrieval, 
QA systems for clarifying inquiries, and calculators for performing mathematical operations, can 
further enhance task performance \cite{abs-2112-00114,abs-2208-03188}. However, 
these methods rely on a single reasoning pathway, if the initial reasoning steps exhibit inaccuracies 
or inherent biases, these defects may propagate through the reasoning process, culminating in erroneous 
conclusions \cite{abs-2212-08073}.

\paragraph{Self-Correction.} 
Self-correction represents a method that allows LLMs to refine their outputs based on feedback 
from prior responses. Two main branches exist for self-correction: fine-tuning and prompting. 
In the field of fine-tuning, critic model \cite{abs-2407-00215} is proposed that assesses 
the responses of LLMs, which is subsequently employed to improve their responses. Conversely, research 
by \cite{abs-2310-20689} and \cite{kumar2024traininglanguagemodelsselfcorrect} recommend
directly fine-tuning the LLM rather than 
training an additional critic model. However, this fine-tuning method demands considerable effort and 
resources.In the field of prompting, research conducted by \cite{KimBM23} 
and \cite{MadaanTGHGW0DPY23} utilizes the LLM's self-reflection and past errors to enhance reasoning. 
PHP \cite{zheng2023progressivehintpromptingimprovesreasoning} employs hints from earlier responses 
to guide models toward more accurate outcomes. Nonetheless, recent studies \cite{0009CMZYSZ24,abs-2407-18219} 
illustrate that even top-tier models frequently struggle with self-correcting reasoning mistakes and 
may experience performance declines without external feedback.
This external feedback may originate from humans, other models \cite{abs-2308-04592,PaulIPBBWF24}, 
or external tools and knowledge sources \cite{GouSGSYDC24,ChenLSZ24,GaoDPCCFZLLJG23}. 
However, in practice, high-quality external feedback is often unavailable. This study proposes a simple 
yet effective external feedback approach that incorporates different perspectives by exchanging answers 
to the same question framed with varied definitions.

\section{Conclusion}

We have introduced the EoP framework, a novel and impactful approach aimed at 
enhancing the reasoning capabilities of LLMs. By integrating external perspectives 
from various definitions of the question, EoP overcomes the limitations of current 
chain-of-thought and self-correction methods, which rely heavily on the model's internal 
perspective. 
This method highlights the significance of comprehensively grasping the question 
prior to developing a solution, and sharing diverse perspectives can cultivate more 
resilient and nuanced reasoning.
Our experimental results across a variety of complex reasoning tasks demonstrate that 
the EoP framework significantly outperforms existing benchmarks, highlighting its 
potential to improve the problem-solving skills of LLMs. In conclusion, our work 
emphasizes the importance of integrating external perspectives obtained from diverse 
problem definitions, setting a new direction for the development of more reliable and 
accurate language models.

\newpage
\appendix

\section{Prompt}
\label{sec:prompt}
\label{sec:redefination-prompt}

\begin{tcolorbox}[breakable, title=PEC Prompt,colback=gray!5,colframe=nmgray!75!black,before skip=1pt, after skip=1pt,fontupper=\linespread{0.8}\selectfont]
{\footnotesize

Extract premises and clear question from input, output a dictionary with 'premise' and 'question' as keys.

\vspace{1em}
\textbf{[Demonstration]}\\
Input:\\
There are 96 fourth-graders at Small Tree School. 43 of them are girls. On Friday, 5 fourth-grade girls and 4 fourth grade boys were absent. How many fourth grade boys were at Small Tree School on Friday? \\
Output:\\
\{\\
"premises":[\\
"Small Tree School has a total of 96 fourth-graders.",\\
"Out of these, 43 are girls.",\\
"On Friday, 5 girls and 4 boys from the fourth grade were absent."\\
],\\
"question": How many fourth-grade boys were present at Small Tree School on Friday?\\
\}

\vspace{1em}
\textbf{[Question to be answered]}\\
Input:\\
\{\{question\}\}\\
Output:\\
\{\\
  "premises": ["string", ...], // all premises extracted from input\\
  "question": string, //core question from input\\
\}\\

}
\end{tcolorbox}

\vspace{1em}
\begin{tcolorbox}[breakable, title=QR Prompt,colback=gray!5,colframe=nmgray!75!black,before skip=1pt, after skip=1pt,fontupper=\linespread{0.8}\selectfont]
    {\footnotesize
    
    Revise and improve the given question while retaining all its original premises and final result:

    \vspace{1em}
    Original question:\\
    \{\{question\}\}
    
    \vspace{1em}
    New question:
    } 
\end{tcolorbox}

\end{document}